%% file: main.tex
\documentclass[10pt,twocolumn,letterpaper]{article}

\usepackage{wacv}              


\definecolor{wacvblue}{rgb}{0.21,0.49,0.74}

\def\wacvPaperID{*****} 
\def\confName{WACV}
\def\confYear{2026}

\title{VReID-XFD: Video-based Person Re-identification at Extreme Far Distance Challenge Results}

\author{
Kailash A. Hambarde$^{1,2}$,
Hugo Proença$^{1,2}$,
Md Rashidunnabi$^{2}$,
Pranita Samale$^{2}$,
Qiwei Yang$^{3}$,
Pingping Zhang$^{3}$, \\
Zijing Gong$^{3}$,
Yuhao Wang$^{3}$,
Xi Zhang$^{3}$,
Ruoshui Qu$^{3}$,
Qiaoyun He$^{3}$,
Yuhang Zhang$^{3}$,
Thi Ngoc Ha Nguyen$^{4}$, \\
Tien-Dung Mai$^{4}$, 
Cheng-Jun Kang$^{5}$,
Yu-Fan Lin$^{5}$,
Jin-Hui Jiang$^{6}$,
Chih-Chung Hsu$^{6}$,
Tamás Endrei$^{7}$,\\
György Cserey$^{7}$,
Ashwat Rajbhandari$^{8}$
\\[0.4em]
$^{1}$IT - Instituto de Telecomunicações, Portugal;
$^{2}$University of Beira Interior, Portugal; \\
$^{3}$Dalian University of Technology, China;
$^{4}$University of Information Technology, VNU-HCM, Vietnam; \\
$^{5}$National Cheng Kung University, Taiwan; 
$^{6}$National Yang Ming Chiao Tung University, Taiwan; \\
$^{7}$ Pázmány Péter Catholic University, Budapest, Hungary;
$^{8}$ Arizona State University, Arizona, USA \\
\\[0.4em]
{\tt\small kailash.hambarde@ubi.pt} (corresponding author)
}

\begin{document}
\maketitle
\input{sec/0_abstract}    
\input{sec/1_intro}
\input{sec/2_related_work}

\input{sec/3_VReID-XFD}

\input{sec/4_discussion}

\input{sec/5_conclusions}
\input{sec/6_appendix}
{
    \small
    \bibliographystyle{ieeenat_fullname}
    \bibliography{main}
}

\end{document}

%% file: sec/0_abstract.tex
\begin{abstract}
Person re-identification (ReID) across aerial and ground at extreme far distances introduces a distinct operating regime where severe resolution degradation, extreme viewpoint changes, unstable motion cues, and clothing variation jointly undermine the appearance-based assumptions of existing ReID systems.
To study this regime, we introduce \textit{VReID-XFD}, a video-based benchmark and community challenge for extreme far-distance (XFD) aerial--ground person re-identification.
VReID-XFD is derived from the DetReIDX dataset and comprises \textit{371 identities, 11{,}288 tracklets, and 11.75 million frames}, captured across altitudes from \textit{5.8\,m to 120\,m}, viewing angles from oblique (30$^\circ$) to nadir (90$^\circ$), and horizontal distances up to 120\,m. The benchmark supports aerial-to-aerial, aerial-to-ground, and ground-to-aerial evaluation under strict identity-disjoint splits, with rich physical metadata.
The VReID-XFD-25 Challenge attracted \textit{10 teams with hundreds of submissions}.
Systematic analysis reveals monotonic performance degradation with altitude and distance, a universal disadvantage of nadir views, and a trade-off between peak performance and robustness.
Even the best performing SAS-PReID method achieves only \textit{43.93\% mAP} in the aerial-to-ground setting. The
dataset, annotations, and official evaluation protocols are publicly available at
\href{https://www.it.ubi.pt/DetReIDX/}{https://www.it.ubi.pt/DetReIDX/}
\end{abstract}

%% file: sec/1_intro.tex
\section{Introduction}
\label{sec:intro}

Person re-identification (ReID) aims to associate individuals across non-overlapping cameras and time \cite{zheng2016person, Nguyen_2024_CVPR}, and is a core component of large-scale surveillance and video analytics. While substantial progress has been achieved in ground-based settings, ReID performance degrades sharply in unconstrained environments due to resolution loss, viewpoint variation, and appearance changes \cite{detreidx, AGReIDv1, nguyen2023ag, nguyen2025ag}.
Recent research has increasingly focused on video-based ReID, exploiting temporal cues such as motion patterns, gait dynamics, and long-term sequence modeling to improve identity discrimination beyond static appearance \cite{rashidunnabi2025causality, hill2025re, nguyen2024attention, zhao2024multi, hambarde2024image}. However, these approaches implicitly assume stable temporal continuity and sufficiently resolved visual evidence assumptions that become fragile when videos are captured at long range under severe physical constraints \cite{nguyen2025person}. Emerging cross-platform and aerial–ground benchmarks further expose these limitations, highlighting the need for robust temporal representations beyond controlled ground-level settings \cite{zhang2024cross, nguyen2025ag, detreidx}.
Unmanned aerial vehicles (UAVs) extend ReID to wide-area monitoring with flexible viewpoints and reduced occlusion \cite{Li2021UAVHumanAL, AGReIDv1}, but introduce a fundamentally more challenging operating regime at extreme far distances. At extended ranges, aerial imagery suffers from severe scale variation, resolution degradation, motion blur, and drastic viewpoint changes; subjects may occupy only a few pixels, rendering fine-grained appearance and motion cues unreliable \cite{detreidx}. In this regime, conventional appearance-driven and temporal aggregation strategies are often mis-specified, leading to systematic performance collapse rather than graceful degradation.
Despite growing interest in aerial ReID, existing benchmarks predominantly focus on aerial-to-aerial or short-range scenarios \cite{Zhang2019PersonRI, Kumar2021ThePA}, leaving \emph{video-based aerial-to-ground ReID at extreme far distances} largely underexplored \cite{detreidx}. Progress has been further constrained by the lack of public benchmarks that jointly capture long-range UAV video, clothing variation, and standardized evaluation protocols.

To address this gap, we introduce the \textit{VReID-XFD 2025 Challenge}, a benchmark and community challenge for video-based person re-identification under extreme far-distance aerial–ground conditions. Built upon the DetReIDX dataset \cite{detreidx} through task-specific filtering and protocol design, VReID-XFD focuses on long-range UAV-to-ground video with severe viewpoint, resolution, and appearance variation. The benchmark provides curated training splits, standardized evaluation metrics, and baseline results, and attracted ten teams with hundreds of submissions. Together, these results expose fundamental failure modes of current video ReID methods and establish a rigorous testbed for studying robustness under extreme physical constraints.

%% file: sec/2_related_work.tex
\section{Related Work}
\label{sec:related}

We review prior work on video-based person re-identification datasets and aerial--ground ReID methods, highlighting limitations that motivate the proposed \textit{VReID-XFD benchmark}.

\subsection{Video-based Person ReID Datasets}

Video-based person re-identification has been widely studied in ground-based surveillance, where temporal information improves identity association across non-overlapping cameras. Early benchmarks such as MARS~\cite{mars} and LS-VID~\cite{GLTR_LS-VID} established large-scale video ReID protocols with thousands of identities and tracklets captured from fixed CCTV cameras, demonstrating the benefits of temporal modeling under relatively stable imaging conditions.
Subsequent datasets emphasized long-term ReID and appearance variation, particularly under clothing changes. CCVID~\cite{gu2022CAL}, VCCR~\cite{han20223d}, and MEVID~\cite{Davila2023mevid} introduced extended temporal spans and cross-clothing scenarios, enabling evaluation of identity persistence over time. Despite their scale and diversity, these datasets remain confined to ground-level viewpoints at close range, where subjects occupy large image regions and viewpoint variation is moderate.
The incorporation of aerial platforms introduces a fundamentally different operating regime. P-Destre~\cite{kumar2020p} represents an early attempt to include UAV video in person ReID, but is limited to low-altitude flights (typically below 10\,m) and short-range observations. More recent efforts, such as the Cross-Platform Video ReID dataset~\cite{zhang2024cross}, explored video-based ReID across aerial and ground cameras at moderate altitudes (20--60\,m), but remain restricted in altitude range, temporal duration, and appearance diversity.
Table~\ref{tab:compare_statics_video} summarizes representative video-based ReID datasets across platforms and operating conditions. As shown, existing benchmarks predominantly focus on ground-only scenarios or limited-altitude aerial views, and do not jointly address long-range UAV video, wide altitude variation, and realistic appearance change. In contrast, VReID-XFD targets aerial-to-ground video across a continuous altitude range of 5--120\,m and different angles with substantially longer sequences and explicit clothing variation, enabling evaluation in a regime underrepresented in prior datasets.

\subsection{Aerial--Ground Person ReID Methods}

Early work on aerial--ground person ReID primarily focused on image-based matching under cross-view appearance discrepancies. The AG-ReID 2023 Challenge~\cite{nguyen2023ag} established standardized protocols for image-level aerial--ground ReID at moderate altitudes, showing that strong ground-based architectures can partially generalize through data augmentation, re-ranking, and cross-view feature alignment.
Most video-based ReID methods have been developed for ground-only settings, relying on temporal aggregation, attention mechanisms, and transformer-based architectures to model motion patterns and appearance consistency~\cite{hou2020temporal, liu2021video, he2021dense, wang2021pyramid}. While effective in controlled surveillance environments, these approaches assume stable temporal continuity and sufficiently resolved visual evidence assumptions that become increasingly unreliable in aerial--ground scenarios due to unstable drone motion, severe resolution degradation, and inconsistent temporal cues.
At extreme far distances, aerial--ground video ReID introduces additional challenges that remain largely unexplored: subjects may occupy only a few pixels, fine-grained appearance cues collapse, and motion information becomes noisy or misleading. Although related work in aerial surveillance and multi-modal recognition provides useful insights~\cite{AerialSurveillance, liu2021watching}, existing methods and benchmarks do not directly address identity association across long-range aerial and ground video streams.
The \emph{VReID-XFD benchmark} bridges this gap by providing a standardized evaluation framework for video-based aerial--ground ReID under extreme far-distance conditions. By combining long-duration UAV video, wide altitude coverage, and realistic appearance variation, \emph{VReID-XFD benchmark} enables systematic analysis of robustness beyond the assumptions of existing benchmarks.

\begin{table}[t]
\caption{Comparison of \emph{VReID-XFD} with representative video-based person ReID datasets.
A/G/W denote Aerial, Ground, and Wearable platforms;
IDs = identities; Trk. = tracklets; Fr. = frames (millions);
Att. = number of annotated attributes; Alt. = operating altitude;
CC = clothing change.}
\label{tab:compare_statics_video}
\centering
\fontsize{7}{7}\selectfont
\setlength{\tabcolsep}{3pt}
\renewcommand{\arraystretch}{1.05}
\begin{tabular}{l c c c c c c c}
\toprule
\textbf{Dataset} &
\textbf{Platform} &
\textbf{IDs} &
\textbf{Trk.} &
\textbf{Fr. (M)} &
\textbf{Att.} &
\textbf{Alt.} &
\textbf{CC} \\
\midrule
MARS~\cite{mars}              
& G   & 1,261 & 20,478 & 1.19 & -- & $<$10\,m & -- \\

LS-VID~\cite{GLTR_LS-VID}     
& G   & 3,772 & 14,943 & 2.98 & -- & $<$10\,m & -- \\

VCCR~\cite{han20223d}         
& G   & 392   & 4,384  & 0.15 & -- & $<$10\,m & \checkmark \\

CCVID~\cite{gu2022CAL}        
& G   & 226   & 2,856  & 0.34 & -- & $<$10\,m & \checkmark \\

MEVID~\cite{Davila2023mevid}  
& G   & 158   & 8,092  & 10.46 & -- & $<$10\,m & \checkmark \\

\midrule
P-Destre~\cite{kumar2020p}    
& A   & 253   & 1,894  & 0.10 & 7  & 5--6\,m  & -- \\

G2A-VReID~\cite{zhang2024cross}
& A/G & 2,788 & 5,576  & 0.18 & -- & 20--60\,m & -- \\

AG-VPReID~\cite{nguyen2025ag}                    
& A/G/W & 3,027 & 13,511 & 3.70 & 15 & 80--120\,m & \checkmark \\

\midrule
\textbf{VReID-XFD}            
& \textbf{A/G} & \textbf{371} & \textbf{11,288} & \textbf{11.75} & \textbf{16} & \textbf{5--120\,m} & \checkmark \\
\bottomrule
\end{tabular}
\end{table}

%% file: sec/3_VReID-XFD.tex
\begin{figure*}[t]
    \centering
    \includegraphics[width=\linewidth]{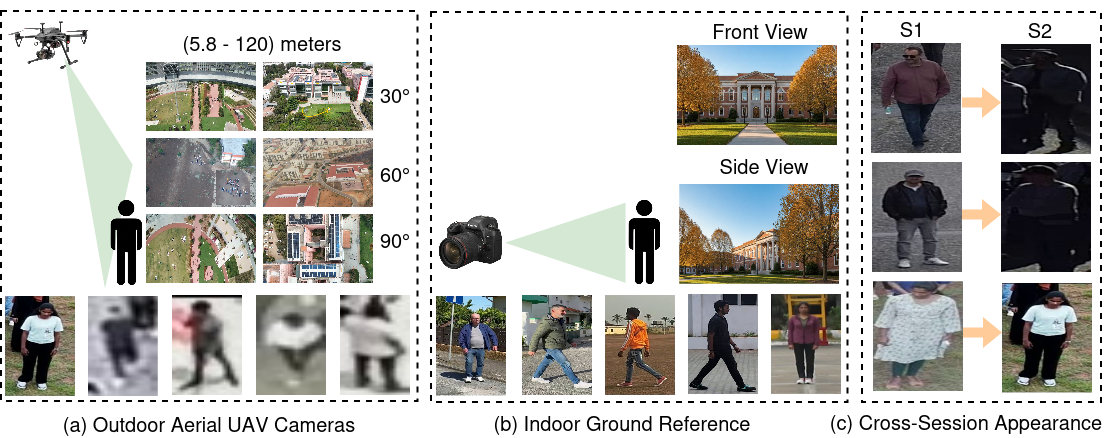}
    \caption{Overview of the VReID-XFD data collection and evaluation protocol. The dataset adopts a two-phase acquisition process with indoor ground reference capture and outdoor UAV-based extreme-distance observation, enabling aerial--ground and aerial--aerial cross-view evaluation.}
    \label{fig:data_protocol}
\end{figure*}

\section{The VReID-XFD 2025 Challenge}
\label{vreidxfd}

\begin{table*}[t]
\caption{VReID-XFD Challenge Results.
\textcolor{darkgreen}{Green} values indicate the best performance,
\textcolor{blue}{blue} values indicate the second-best performance, and
\textcolor{red}{red} values indicate the third-best performance for each test protocol and metric.}
\label{tab:vreidxfd-results}
\centering
\renewcommand{\tabcolsep}{4pt}
\begin{tabular}{c||c|c|c|c||c|c|c|c||c|c|c|c}
\toprule
\multirow{2}{*}{\textbf{Method}} 
& \multicolumn{4}{c||}{\textbf{Aerial$\rightarrow$Aerial}} 
& \multicolumn{4}{c||}{\textbf{Aerial$\rightarrow$Ground}} 
& \multicolumn{4}{c}{\textbf{Ground$\rightarrow$Aerial}} \\
\cline{2-13}
& R1 & R5 & R10 & mAP
& R1 & R5 & R10 & mAP
& R1 & R5 & R10 & mAP \\
\midrule

VSLA \cite{zhang2024cross}
& 15.96 & 26.10 & 32.77 & 13.83
& 28.96 & 54.71 & 69.13 & \textcolor{blue}{41.63}
& 58.43 & 65.17 & 69.66 & 26.26 \\

SINet \cite{bai2022salient}
& 14.06 & 24.51 & 30.94 & 12.85
& 25.62 & 52.47 & 66.57 & 38.46
& 23.50 & 49.44 & 59.55 & 16.98 \\

PSTA \cite{wang2021pyramid}
& 13.00 & 23.80 & 30.30 & 10.50
& 22.30 & 46.90 & 59.70 & 34.40
& 40.40 & 56.20 & 59.60 & 17.00 \\

BiCNet-TKS \cite{hou2021bicnet}
& 13.30 & 26.78 & 36.57 & 9.71
& 21.71 & 44.85 & 59.30 & 33.28
& 41.57 & 58.43 & 65.17 & 22.12 \\

\hline

DUT\_IIAU\_LAB
& \textcolor{darkgreen}{25.39} & \textcolor{darkgreen}{39.58} & \textcolor{darkgreen}{48.44} & \textcolor{darkgreen}{20.13}
& \textcolor{darkgreen}{37.77} & \textcolor{darkgreen}{65.26} & \textcolor{darkgreen}{75.31} & \textcolor{darkgreen}{43.93}
& \textcolor{darkgreen}{69.66} & \textcolor{darkgreen}{80.90} & \textcolor{darkgreen}{87.64} & \textcolor{darkgreen}{35.44} \\

H Nguyn\_UIT
& \textcolor{blue}{20.37} & 29.08 & 34.85 & \textcolor{blue}{19.79}
& \textcolor{red}{33.15} & \textcolor{blue}{60.49} & \textcolor{blue}{74.88} & 39.59
& \textcolor{blue}{62.92} & 66.29 & 69.66 & \textcolor{blue}{34.49} \\

CJKang
& \textcolor{red}{20.08} & \textcolor{blue}{31.19} & \textcolor{blue}{38.54} & \textcolor{red}{16.81}
& \textcolor{blue}{33.65} & 57.68 & 73.38 & \textcolor{red}{39.63}
& \textcolor{blue}{62.92} & \textcolor{red}{69.66} & \textcolor{red}{75.28} & 28.59 \\

PPCUITK
& 19.15 & \textcolor{red}{30.30} & \textcolor{red}{36.86} & 16.66
& 32.77 & \textcolor{red}{59.64} & 71.97 & 39.21
& \textcolor{blue}{62.92} & \textcolor{red}{69.66} & 73.03 & \textcolor{red}{29.75} \\

JNNCE ISE
& 19.00 & 30.05 & 36.82 & 16.52
& 32.05 & 59.21 & 73.06 & 38.50
& \textcolor{red}{64.04} & 68.54 & 70.79 & 29.48 \\

Rajbhandari Ashwat ASU
& 18.68 & 29.97 & 38.29 & 16.02
& 32.37 & 59.43 & \textcolor{red}{74.75} & 38.76
& \textcolor{blue}{62.92} & \textcolor{blue}{75.28} & \textcolor{blue}{77.53} & 28.36 \\

Yu Fan Lin
& 11.04 & 20.44 & 27.00 & 9.80
& 22.85 & 48.16 & 62.80 & 29.17
& 26.97 & 44.94 & 52.81 & 14.47 \\

\bottomrule
\end{tabular}
\end{table*}

\subsection{Dataset}

The \emph{VReID-XFD 2025 Challenge} is built upon the \emph{DetReIDX} dataset~\cite{detreidx}, comprising 371 identities, 11,288 tracklets, and approximately 11.75 million frames. Data were collected across seven university campuses in Portugal, Turkey, Angola, and India, ensuring substantial diversity in subject appearance, clothing styles, environments, and demographics. This geographic diversity supports realistic and large-scale evaluation of cross-view person re-identification under unconstrained conditions.

\begin{figure}[t]
    \centering
    \includegraphics[width=\linewidth, height=6cm]{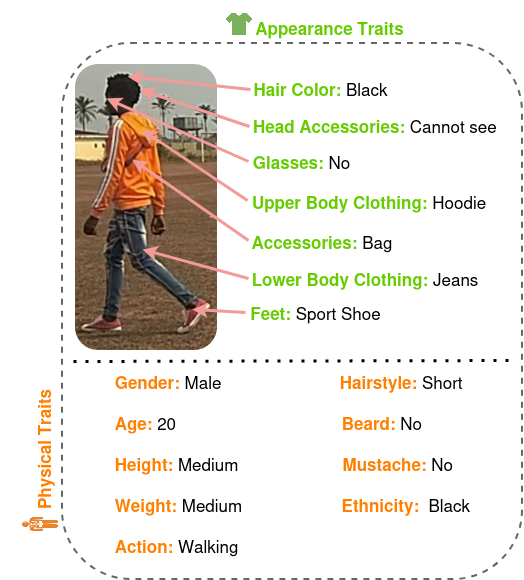}
    \caption{Distribution of annotated soft-biometric attributes in VReID-XFD, including appearance, clothing, and physical traits used for analysis and evaluation.}
    \label{fig:softbio}
\end{figure}

\begin{figure}[t]
    \centering
    \includegraphics[width=\linewidth, height=6cm]{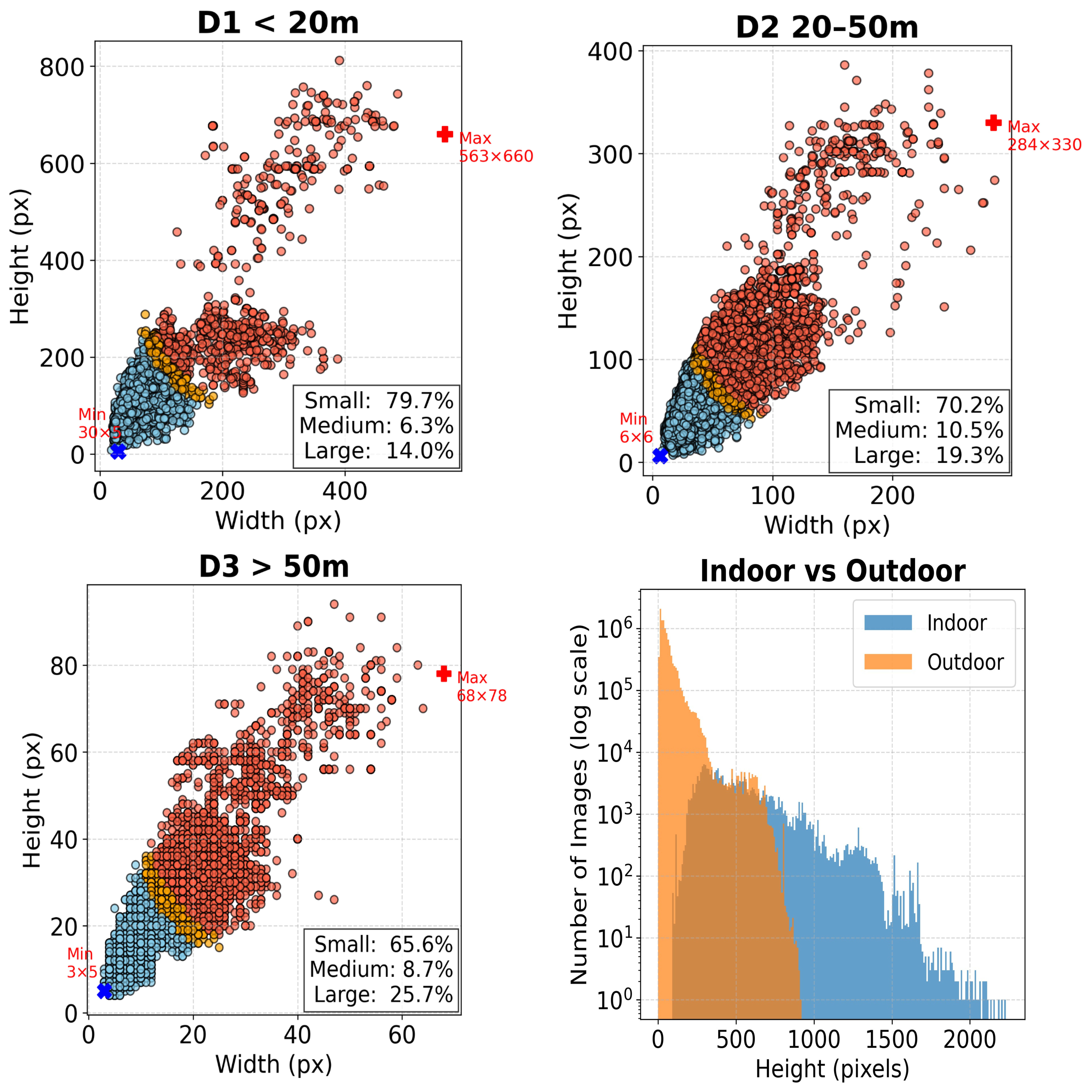}
    \caption{Scatter plots of ROIs height/width in three different distance bins. The bottom-right plot compares the distribution of the ROIs height (in pixels) in indoor and outdoor data.}
    \label{fig:height_width_distribution}
\end{figure}

\subsubsection{Two-Phase Collection Protocol}

DetReIDX adopts a two-phase acquisition protocol designed to decouple identity reference capture from extreme-distance aerial observation.

\textbf{Ground Reference.}
Each subject is first recorded using DSLR and smartphone cameras, capturing a 20-second walking sequence with natural posture variation. These sequences provide stable identity references under controlled imaging conditions.

\textbf{Outdoor UAV Capture.}
Subjects are then recorded outdoors in two independent sessions (S1 and S2), wearing different outfits to introduce appearance variation. For each session, 18 UAV viewpoints are captured by varying pitch angle (30°, 60°, 90°), altitude (5.8–120\,m), and horizontal distance (10–120\,m). This design induces systematic variation in scale, resolution, and viewpoint, enabling analysis of performance degradation under extreme far-distance conditions Human body sizes vary dramatically from for UAV imagery, posing unprecedented challenges, as illustrated in Fig. \ref{fig:height_width_distribution}.
The dataset includes 16 soft-biometric labels such as gender, age, clothing, and accessories, as shown in Fig. \ref{fig:softbio}.

\subsubsection{Dataset Splits and Evaluation Protocols}

We define three evaluation protocols following standard cross-view ReID settings:
\begin{itemize}
    \item \textbf{Aerial$\rightarrow$Aerial (A$\rightarrow$A):} Matching UAV tracklets across disjoint aerial viewpoints.
    \item \textbf{Aerial$\rightarrow$Ground (A$\rightarrow$G):} UAV queries matched against ground-based gallery tracklets.
    \item \textbf{Ground$\rightarrow$Aerial (G$\rightarrow$A):} Ground-based queries matched against UAV gallery tracklets.
\end{itemize}
All test identities are strictly disjoint from training identities.

\subsection{Submission and Participation}

The challenge was hosted on \emph{Kaggle}\footnote{https://www.kaggle.com/competitions/detreidxv1/} with a standardized evaluation server. Participants submitted ranked gallery tracklet predictions following the DetReIDX protocol~\cite{detreidx} and final rankings were determined by overall mAP averaged across all three protocols.
The challenge attracted 10 teams from international research institutions, generating hundreds of submissions. Six teams surpassed the VSLA baseline~\cite{zhang2024cross}, indicating meaningful progress despite the task’s difficulty.

\subsection{Evaluation}

The quantitative results of the \emph{Kaggle VReID-XFD 2025 Challenge} \footnote{https://www.kaggle.com/competitions/detreidxv1/leaderboard} are summarized in Table~\ref{tab:vreidxfd-results}, reporting Rank-1, Rank-5, Rank-10 accuracy, and mean Average Precision (mAP) across the A$\rightarrow$A, A$\rightarrow$G, and G$\rightarrow$A protocols.
Across all three protocols, \emph{DUT\_IIAU\_LAB} consistently achieves the best performance, ranking first for every reported metric. In the most challenging A$\rightarrow$G setting, it attains 37.77\% Rank-1 accuracy and 43.93\% mAP, clearly outperforming all other submissions and establishing a strong benchmark for extreme far-distance aerial--ground ReID. Similar dominance is observed in the A$\rightarrow$A protocol, where DUT\_IIAU\_LAB achieves 25.39\% Rank-1 and 20.13\% mAP, as well as in the G$\rightarrow$A protocol, where it reaches 69.66\% Rank-1 and 35.44\% mAP.
For second-best performance, \emph{H~Nguyn\_UIT} consistently ranks immediately behind the top method across most metrics and protocols. It achieves 20.37\% Rank-1 and 19.79\% mAP in A$\rightarrow$A, 33.15\% Rank-1 and 39.59\% mAP in A$\rightarrow$G, and 62.92\% Rank-1 with 34.49\% mAP in G$\rightarrow$A. These results indicate strong generalization across viewpoints, particularly under cross-view aerial--ground matching.
The third-best performance is shared across multiple competitive teams depending on the protocol and metric. In A$\rightarrow$A, \emph{CJKang} ranks third with 20.08\% Rank-1 and 16.81\% mAP, closely followed by \emph{PPCUITK} and \emph{JNNCE~ISE}. In the A$\rightarrow$G protocol, \emph{CJKang} again ranks third with 33.65\% Rank-1 and 39.63\% mAP, while \emph{PPCUITK} and \emph{Rajbhandari~Ashwat~ASU} achieve comparable performance. For G$\rightarrow$A, \emph{JNNCE~ISE} attains the third-highest Rank-1 accuracy (64.04\%), whereas \emph{Rajbhandari~Ashwat~ASU} ranks third in Rank-5 and Rank-10 accuracy, reflecting strong retrieval quality at higher ranks.
All reported baselines approaches, including \emph{SINet}, \emph{PSTA}, \emph{BiCNet-TKS}, and \emph{Yu~Fan~Lin}, exhibit lower overall performance, particularly in the A$\rightarrow$A protocol. This highlights the intrinsic difficulty of bilateral extreme-distance matching when both query and gallery sequences suffer from severe resolution degradation.

\begin{figure}
    \centering
    \includegraphics[width=\linewidth]{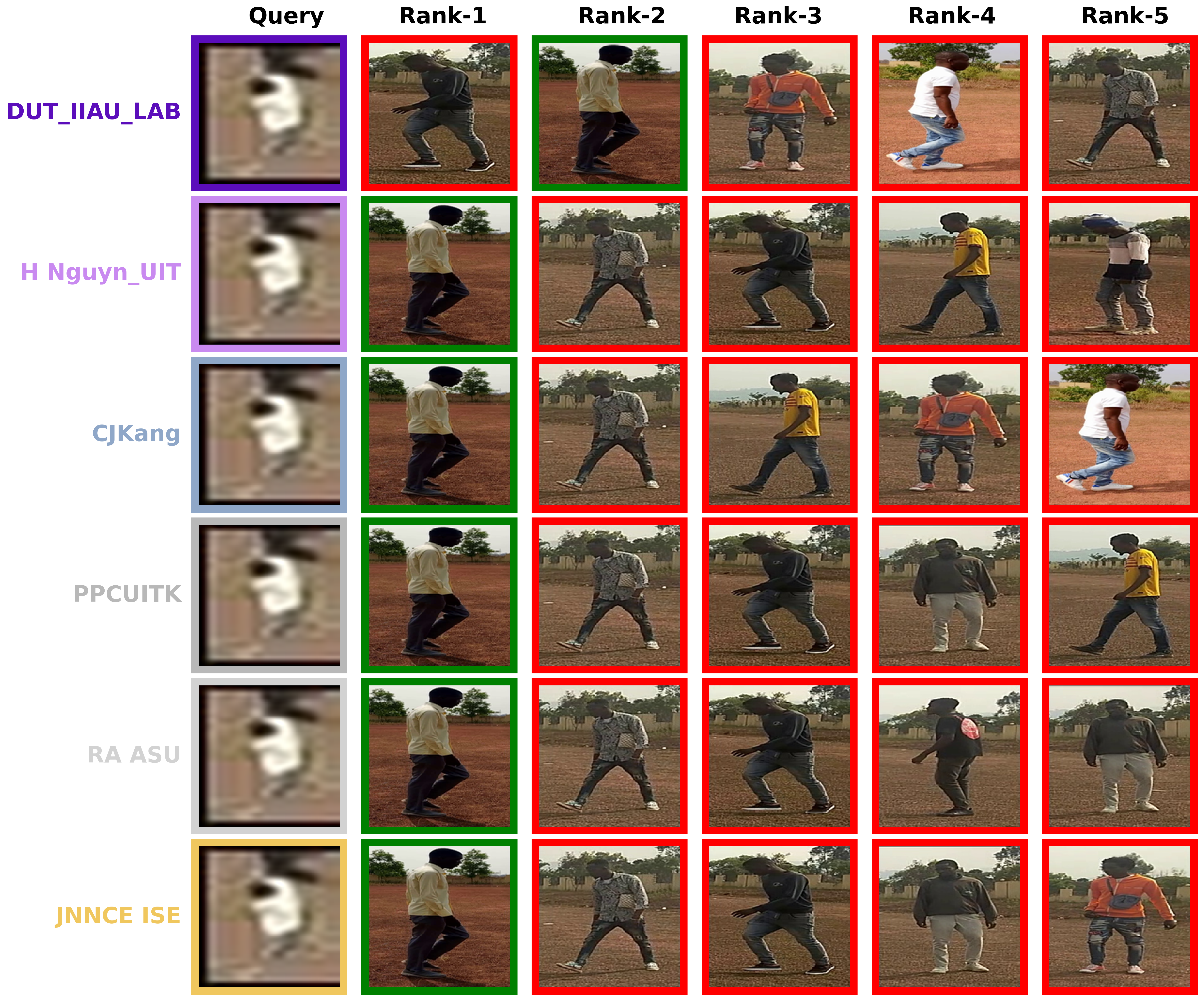}
    \caption{Aerial→Ground test case at high altitude and top angle. Green/red: correct/incorrect labels.}
    \label{fig:quanti_a2g}
\end{figure}

%% file: sec/4_discussion.tex
\begin{figure*}[t]
    \centering

    \begin{subfigure}[t]{0.48\textwidth}
        \centering
        \includegraphics[width=\linewidth]{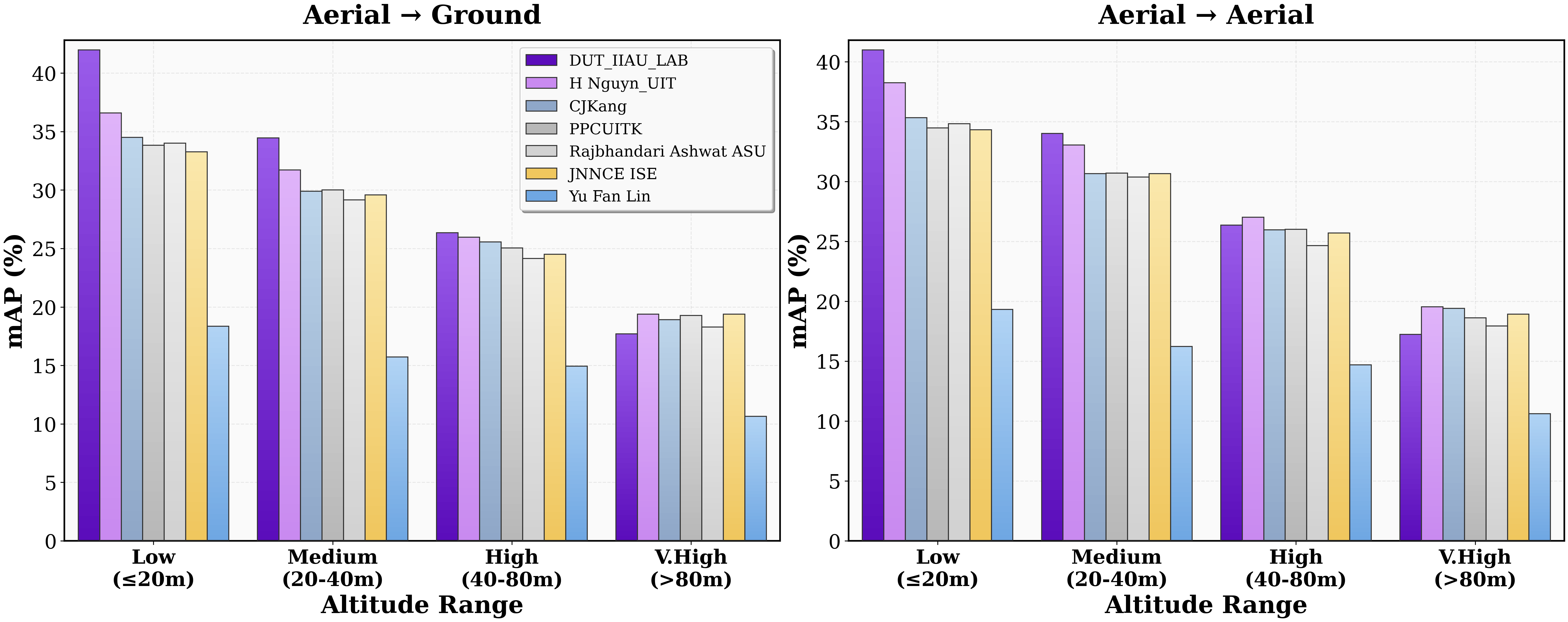}
        \caption{Altitude-specific performance analysis.}
        \label{fig:discussion_altitude}
    \end{subfigure}
    \hfill
    \begin{subfigure}[t]{0.48\textwidth}
        \centering
        \includegraphics[width=\linewidth]{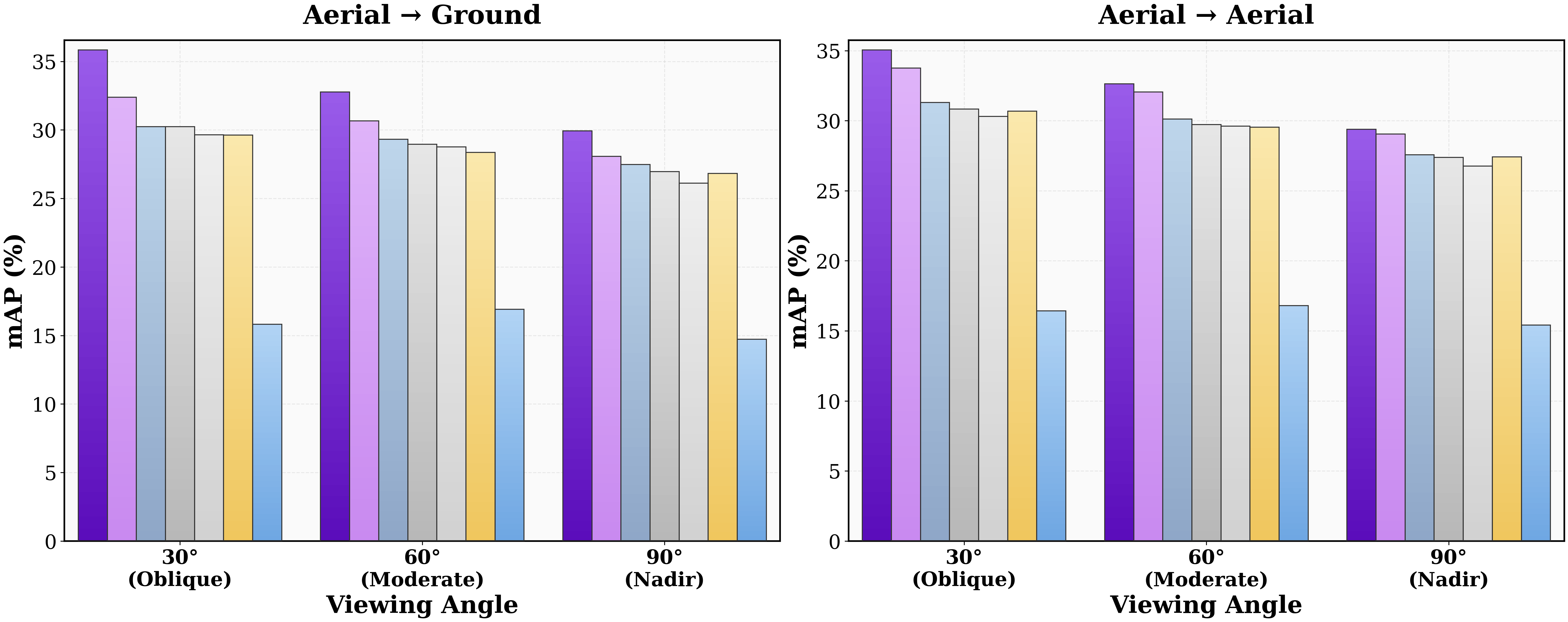}
        \caption{Angle-specific performance analysis.}
        \label{fig:discussion_angle}
    \end{subfigure}

    \vspace{0.5em}

    \begin{subfigure}[t]{0.6\textwidth}
        \centering
        \includegraphics[width=\linewidth]{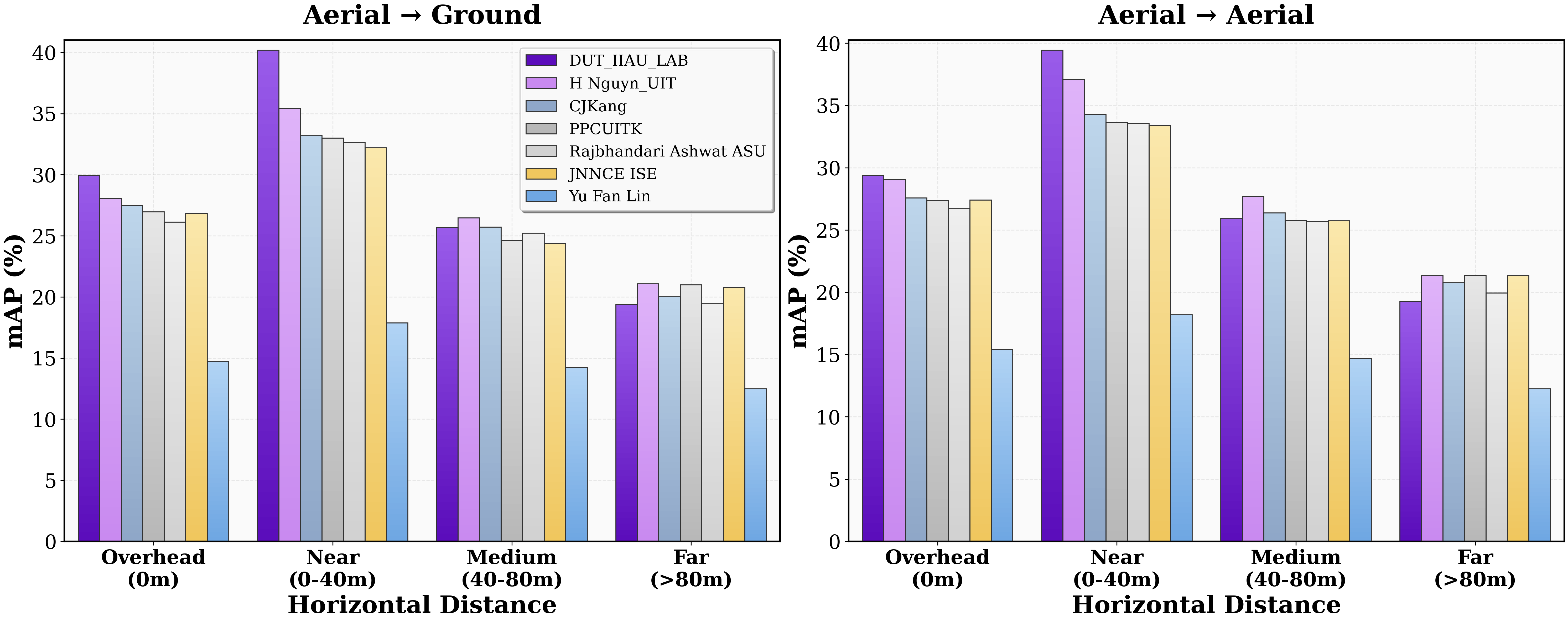}
        \caption{Distance-specific performance analysis.}
        \label{fig:discussion_distance}
    \end{subfigure}

    \caption{Comprehensive analysis of physical factors affecting video-based person re-identification performance under extreme far-distance conditions.
    (a) Effect of altitude, (b) effect of viewing angle, and (c) effect of horizontal distance.
    All methods exhibit monotonic degradation under increasingly adverse imaging conditions.}
    \label{fig:discussion_comprehensive}
\end{figure*}

\section{Discussion}
\label{discussion}

The \emph{VReID-XFD benchmark} provides explicit physical metadata that enables a systematic analysis of how altitude, viewing angle, and horizontal distance influence video-based person re-identification performance under extreme far-distance (XFD) conditions. Figure~\ref{fig:discussion_comprehensive} consolidates these effects and reveals consistent performance trends across all participating methods, exposing fundamental limitations of current ReID systems when deployed in realistic aerial surveillance scenarios.

\subsection{Altitude-Induced Scale Degradation}

The effect of altitude is illustrated in Fig.~\ref{fig:discussion_comprehensive}(a) and Fig. ~\ref{fig:quanti_a2g}. Across all teams and both evaluation protocols (A$\rightarrow$G and A$\rightarrow$A), performance decreases monotonically as altitude increases from low to very high ranges. This trend is consistent across methods and represents the most significant source of performance degradation among all physical factors considered.
At low altitude, where subjects occupy relatively large spatial extents, methods achieve their highest mAP values. As altitude increases beyond 80\,m, performance drops sharply. For A$\rightarrow$G, the average mAP decreases from 33.22\% at low altitude to 17.66\% at very high altitude, while for A$\rightarrow$A the degradation is even more severe, falling from 23.11\% to 6.64\%. Although DUT\_IIAU\_LAB achieves the strongest performance at low altitude, its accuracy declines substantially at extreme heights. In contrast, H~Nguyn\_UIT and JNNCE~ISE maintain the highest absolute performance at very high altitude (19.39\% mAP for A$\rightarrow$G), indicating relatively better robustness under severe scale degradation.
This behavior closely reflects changes in ground sampling distance. At extreme altitudes, a person occupies only a small number of pixels in height, rendering fine-grained appearance cues such as clothing texture, facial details, and accessories effectively unobservable. Under these conditions, recognition relies primarily on coarse body shape and global color statistics, which are inherently less discriminative. Consequently, the task shifts from conventional appearance-based ReID to recognition under extreme information loss.

\subsection{Effect of Viewing Angle and Geometric Distortion}

The influence of viewing angle is shown in Fig.~\ref{fig:discussion_comprehensive}(b). All methods follow a consistent ordering across angles: oblique views (30°) yield the highest performance, followed by moderate angles (60°), while nadir views (90°) are the most challenging. Although the magnitude of angle-induced degradation is smaller than that caused by altitude, it is systematic across all teams.
For A$\rightarrow$G, average mAP decreases from 29.11\% at 30° to 25.73\% at 90°, corresponding to a 3.38\% drop. A similar trend is observed for A$\rightarrow$A, where performance declines from 29.76\% to 26.14\%. DUT\_IIAU\_LAB achieves the highest accuracy at all angles but also exhibits the largest oblique-to-nadir performance gap (5.91\%), whereas Yu~Fan~Lin shows minimal sensitivity to viewing angle at the cost of lower overall accuracy.
The observed trade-off between peak accuracy and angular robustness suggests that learning representations that are both highly discriminative and strongly viewpoint-invariant remains a challenging open problem.

\subsection{Influence of Horizontal Distance and Long-Range Effects}

The effect of horizontal distance is presented in Fig.~\ref{fig:discussion_comprehensive}(c). For both A$\rightarrow$G and A$\rightarrow$A protocols, far-range queries (greater than 80\,m) are consistently the most challenging, with average mAP values of 19.17\% and 19.46\%, respectively. Near-range observations achieve the highest performance, followed by overhead and medium-distance cases.
An interesting observation is the moderate performance achieved in the overhead setting despite pure nadir viewing. This suggests that higher effective resolution at zero horizontal distance can partially compensate for geometric distortion. 
\subsection{Performance Limits and Research Implications}

Despite consistent improvements over baseline methods, the analysis in Fig.~\ref{fig:discussion_comprehensive} indicates a substantial gap between current automated systems and reliable recognition under compound adverse conditions, particularly when very high altitude, nadir viewing, and far-range capture occur simultaneously. Under such worst case combinations, expected performance drops to approximately 10--15\% mAP, approaching random retrieval for large galleries.

The results also reveal that different methods exhibit distinct trade-offs between peak accuracy under favorable conditions and stability under extreme degradation. While DUT\_IIAU\_LAB achieves the strongest overall performance, other approaches demonstrate comparatively better robustness at extreme altitudes, underscoring the need to balance discriminative power with resilience to physical degradation.

%% file: sec/5_conclusions.tex
\section{Conclusion}

This paper presents the results of the \emph{VReID-XFD 2025 Challenge}, a large-scale evaluation of video-based person re-identification under extreme far-distance aerial–ground conditions. The benchmark targets a highly challenging regime characterized by severe altitude variation, extreme viewpoint changes, long-range capture, and significant resolution degradation.
Across all submissions, DUT\_IIAU\_LAB achieved the strongest overall performance, reaching 43.93\% mAP and 37.77\% Rank-1 in the most challenging Aerial-to-Ground setting. H Nguyn\_UIT and JNNCE ISE demonstrated more stable performance at very high altitudes, while other teams consistently outperformed the VSLA baseline across multiple protocols. Overall, six teams surpassed the baseline, indicating steady progress despite the difficulty of the task.
Analysis of the challenge results reveals monotonic performance degradation with increasing altitude and distance, consistent underperformance of nadir viewpoints compared to oblique views, and a clear advantage of Ground-to-Aerial matching over Aerial-to-Ground scenarios. Under worst-case conditions combining very high altitude, nadir viewing, and far-range capture, performance drops to approximately 10–15\% mAP, highlighting the fundamental limitations of current appearance-based methods.
The VReID-XFD benchmark establishes a rigorous evaluation platform for extreme-distance video-based person re-identification and provides a foundation for future research toward more robust and operationally reliable aerial surveillance systems.

%% file: sec/6_appendix.tex
\appendix
\section{Submitted Person ReID Algorithms}
\label{appendix}

In the following appendix, we provide technical summaries of the video-based person ReID algorithms assessed during the VReID-XFD2025 Challenge. Teams are listed in order of final leaderboard ranking. Detailed technical reports were provided by participating teams.

\subsection{SAS-VPReID: A Scale-Adaptive Framework with Shape Priors for Video-based Person Re-Identification at Extreme Far Distances}
\textit{Qiwei Yang, Pingping Zhang Zijing Gong, Yuhao Wang, Xi Zhang, Ruoshui Qu, Qiaoyun He, Yuhang Zhang}

\textbf{Description of the Algorithm:}

The VReID-XFD challenge addresses video-based person re-identification under extreme far-distance (XFD) conditions, characterized by severe aerial–ground viewpoint gaps, strong cross-session appearance variations, and weak, noisy, and temporally unstable frame-level cues.

\begin{figure}[h]
    \centering
    \includegraphics[width=0.95\linewidth]{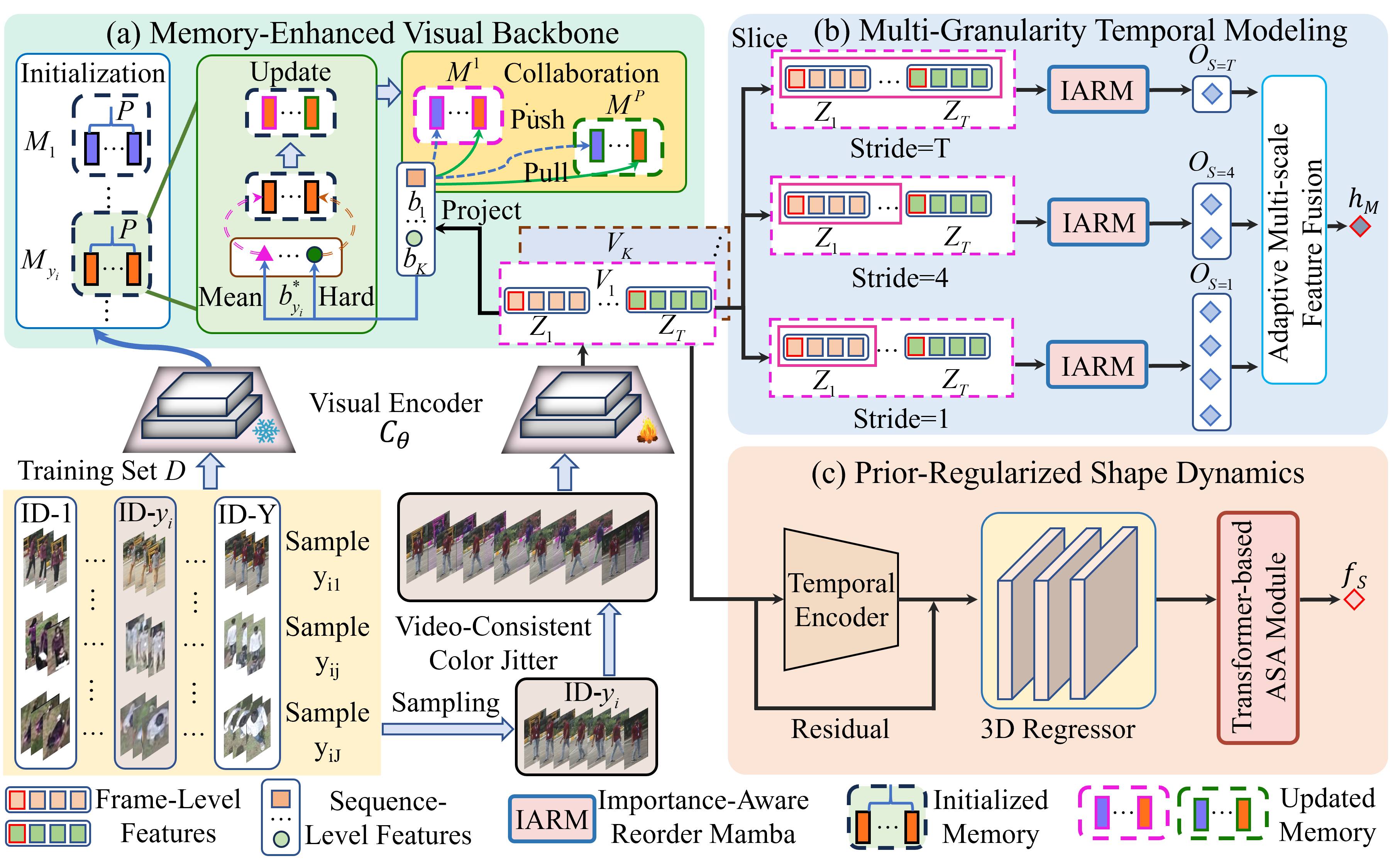}
    \caption{DUT\_IIAU\_LAB: Vision Transformer-based framework with temporal aggregation and cross-view feature learning.}
    \label{fig:dut_framework}
\end{figure}

\paragraph{Method.}
As shown in fig. \ref{fig:dut_framework} we propose \textbf{SAS-VPReID}, a scale-adaptive and shape-aware framework designed to improve robustness to extreme scale variation, viewpoint changes, temporal noise, and clothing-induced cross-session mismatch. The framework comprises three components:

\begin{itemize}
    \item \textbf{MEVB: Memory-Enhanced Visual Backbone.}
    CLIP ViT-L/14 is adopted to enhance representation for low-resolution XFD videos. Video-Consistent Color Jitter preserves temporal coherence across frames, while a multi-proxy momentum memory stabilizes identity learning under viewpoint, blur, clothing, and illumination variations.

    \item \textbf{MGTM: Multi-Granularity Temporal Modeling.}
    Tracklets are sliced at multiple temporal granularities and encoded using Mamba-style temporal encoders. A learnable fusion adaptively weights temporal scales per instance, emphasizing the most informative dynamics.

    \item \textbf{PRSD: Prior-Regularized Shape Dynamics.}
    Per-frame 3D shape parameters are regressed and modeled temporally as clothing-insensitive cues. An explicit SMPL shape prior regularizes shape estimation under blur and occlusion, complementing appearance features for cross-session robustness.
\end{itemize}

\subsection{Enhancing Aerial--Ground Video Person Re-Identification via DFGS-Guided CLIP Sampling and Inference-Time Uncertainty-Aware Fusion}
\textit{Thi Ngoc Ha Nguyen, Tien-Dung Mai}

\begin{figure}[h]
    \centering
    \includegraphics[width=0.95\linewidth]{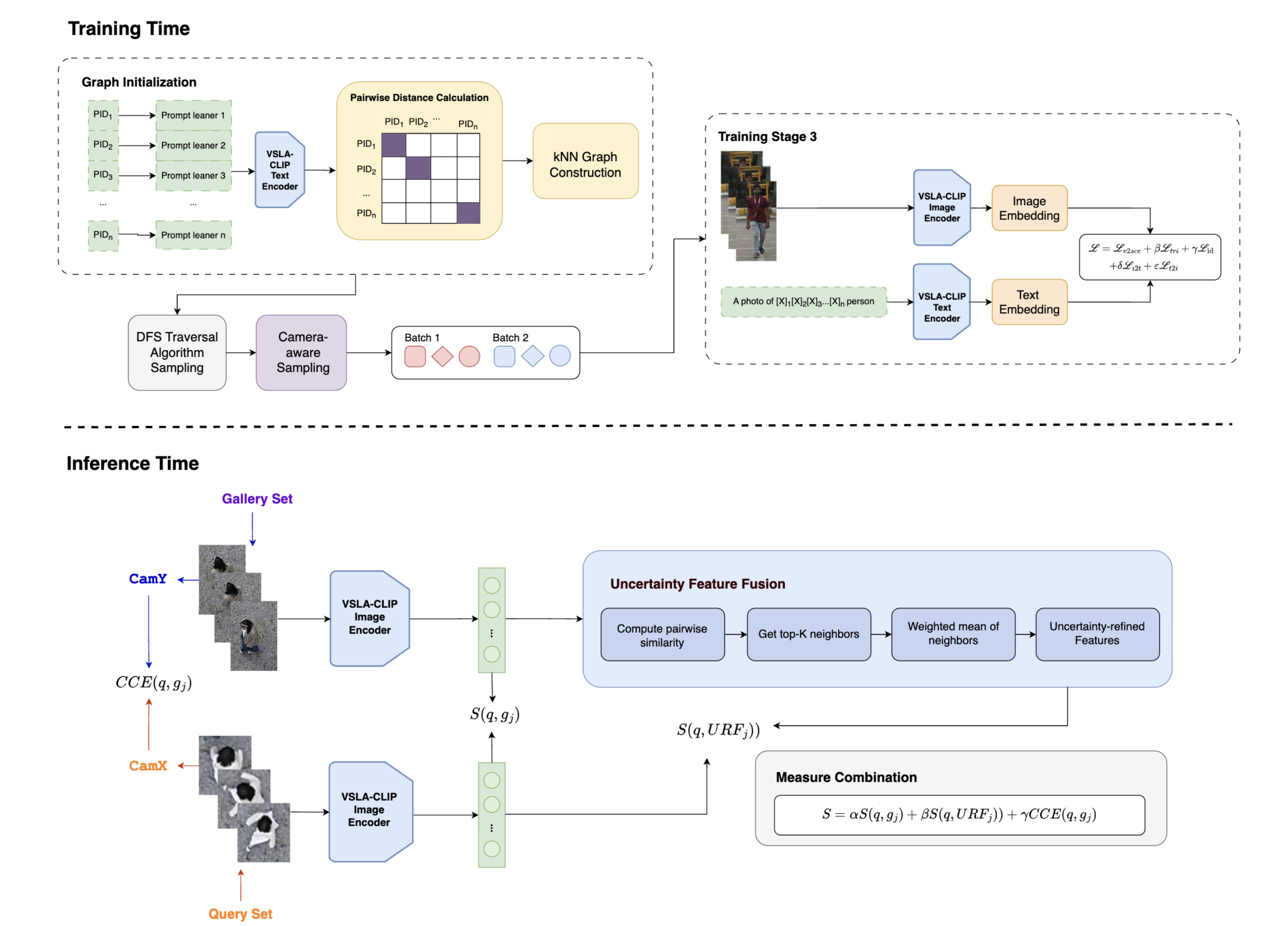}
    \caption{H~Nguyen\_UIT: CLIP-based framework with training-stage sampling refinement and inference-time uncertainty-aware fusion.}
    \label{fig:hnguyen_framework}
\end{figure}

\subsubsection{Description of the Algorithm}

This method presents a competition-oriented solution for aerial--ground video-based person re-identification on the DetReIDX benchmark. The approach builds upon the VSLA-CLIP framework~\cite{zhang2024cross} and improves training efficiency and inference robustness without modifying the backbone architecture or introducing additional supervision.

\paragraph{Training Strategy.}
To mitigate performance saturation and mild overfitting, Stage-2 training is shortened to 60 epochs while keeping all other training settings unchanged. During Stage-3 fine-tuning, a Depth-First Graph Sampler (DFGS)~\cite{zhao2024clip} is introduced to emphasize semantically ambiguous identities, without altering the network architecture or optimization objectives.

\paragraph{Inference-Time Enhancement.}
To reduce noise in gallery representations, an Uncertainty Feature Fusion Method (UFFM)~\cite{che2025enhancing} is applied to aggregate local neighborhood features in a training-free manner. In addition, Camera Consistency Encoding (CCE) is incorporated as a weak camera-aware prior. Multiple similarity cues are combined at inference time using fixed weights:
\begin{equation}
S^{*}(q, g_j) = \alpha S(q, g_j) + \beta S(q, \mathrm{URF}_j) + \gamma \mathrm{CCE}(q, g_j),
\end{equation}
where $\alpha + \beta + \gamma = 1$.

\paragraph{Summary.}
The proposed solution enhances VSLA-CLIP through (i) reduced Stage-2 training duration, (ii) DFGS-based fine-tuning in Stage-3, and (iii) training-free inference improvements using UFFM, CCE, and multi-measure similarity fusion.

\subsection{EAGLE-ReID: Strategic Alignment and Delta Consistency for Extreme Far-Distance Aerial--Ground Re-Identification}
\textit{Cheng-Jun Kang, Jin-Hui Jiang, Yu-Fan Lin, Chih-Chung Hsu}

\begin{figure}[h]
    \centering
    \includegraphics[width=0.95\linewidth]{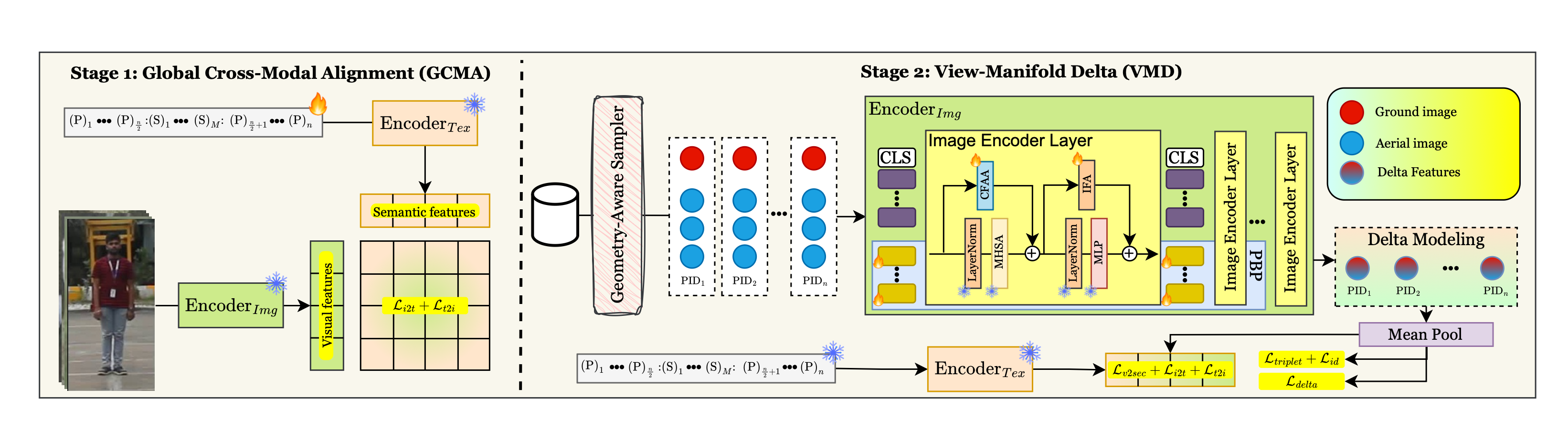}
    \caption{EAGLE-ReID: Framework overview illustrating Geometry-Aware Sampling and View-Manifold Delta modeling for cross-view alignment.}
    \label{fig:eagle_framework}
\end{figure}

\subsubsection{Description of the Algorithm}

The VReID-XFD challenge presents an extreme far-distance (XFD) scenario where the domain gap between ground-view and high-altitude aerial-view videos is severe. Preliminary analysis reveals two critical limitations in standard vision--language model baselines: (i) data imbalance, where scarce ground samples are overwhelmed by abundant aerial samples during training, and (ii) manifold disjointness, where ground and aerial feature distributions remain separated even after fine-tuning.

\paragraph{Method.}
To address these challenges, the authors propose \textbf{EAGLE-ReID}, a framework built upon the VSLA-CLIP architecture with a frozen ViT-B/16 backbone and learnable adapters. The method focuses on enforcing strategic cross-view alignment through targeted training strategies, rather than modifying the backbone or introducing additional supervision. Two key components are introduced:

\begin{itemize}
    \item \textbf{Geometry-Aware Sampler (GAS).}
    Standard PK-sampling often fails to construct valid cross-view pairs due to the imbalance between ground and aerial samples. GAS enforces a structured batch composition by selecting exactly one ground anchor and $K$ aerial instances per identity, ensuring that each optimization step contains valid ground--aerial pairs and effectively activates cross-view triplet supervision.

    \item \textbf{View-Manifold Delta (VMD).}
    Instead of directly aligning features across the extreme domain gap, VMD models discriminative transformations in a difference space. For each aerial instance, a delta vector is defined as the difference between aerial and ground features. A Delta Triplet Loss enforces consistency of ground-to-aerial transformations for the same identity while maintaining discriminability across identities. An additional intra-aerial consistency loss further enhances discrimination within the aerial domain.
\end{itemize}

\subsection{S3-CLIP: Video Super-Resolution for Person Re-Identification}
\textit{Tam\'as Endrei, Gy\"orgy Cserey}

\begin{figure}[h]
    \centering
    \includegraphics[width=0.95\linewidth]{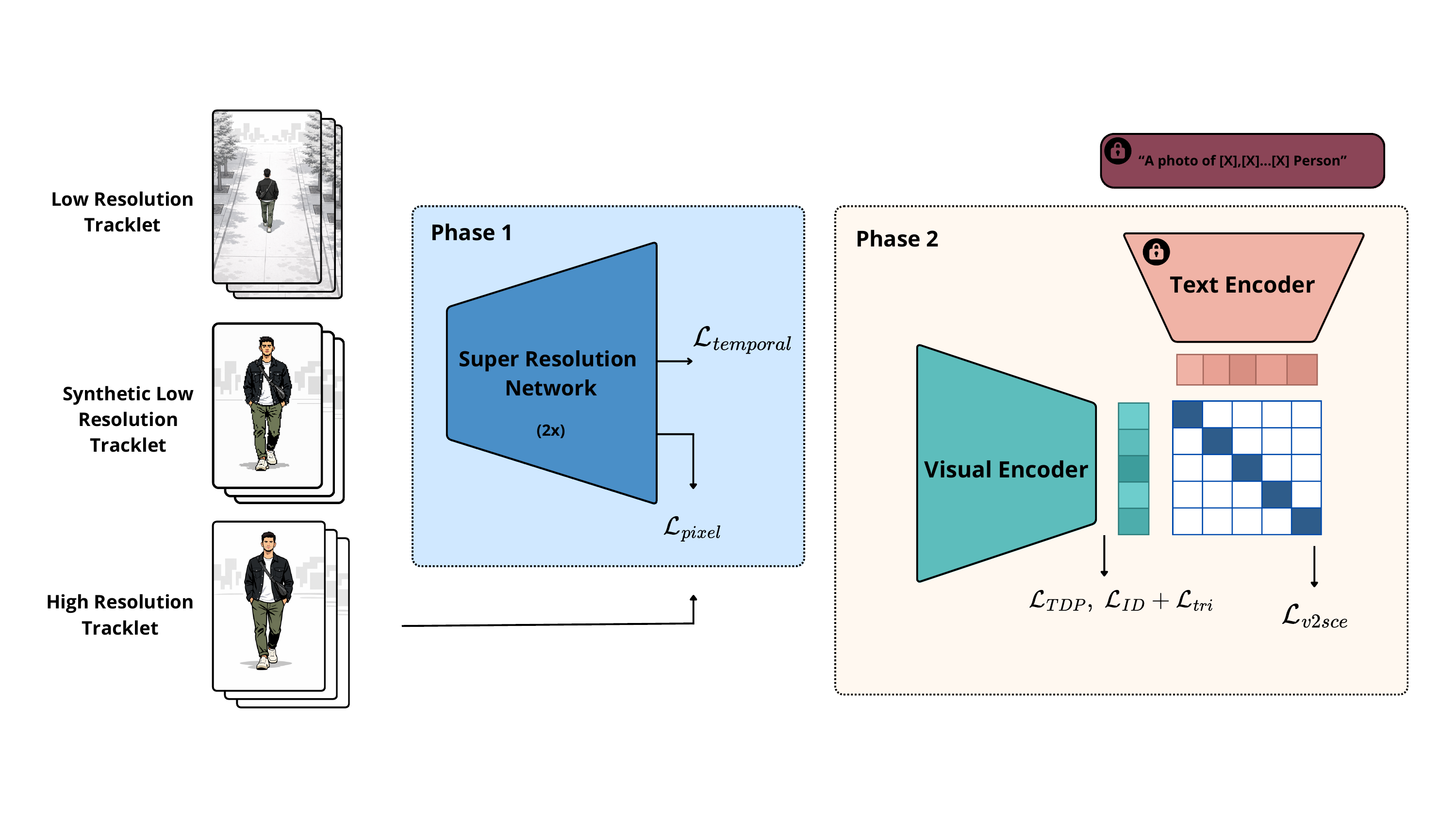}
    \caption{S3-CLIP: Two-phase training strategy integrating a super-resolution module with a CLIP-based visual encoder for low-resolution video ReID.}
    \label{fig:s3clip_framework}
\end{figure}

\subsubsection{Description of the Algorithm}

S3-CLIP addresses video-based person re-identification under extreme far-distance conditions by explicitly enhancing low-resolution visual evidence through video super-resolution. The method extends the VSLA-CLIP framework by integrating a super-resolution (SR) module as a preprocessing stage, while remaining backbone-agnostic and not requiring additional supervision.

\paragraph{Method.}
The proposed framework incorporates three key elements from recent super-resolution literature: SING-based sampling to mitigate the scarcity of paired high- and low-resolution tracklets, SwinIR for frame-level super-resolution, and task-driven perceptual supervision to align SR optimization with the downstream ReID objective. Tracklets are categorized based on a predefined resolution threshold; high-resolution tracklets are synthetically downscaled to generate paired low-resolution samples, which are concatenated with naturally low-resolution tracklets and processed by the SR network.

A two-phase training strategy is adopted to ensure stable optimization. In the first phase, only the super-resolution network is optimized while the CLIP-based visual encoder remains frozen. In the second phase, the SR network is frozen and the visual encoder is fine-tuned, preventing early overfitting to SR-induced artifacts. The overall optimization combines super-resolution losses, including pixel reconstruction, temporal consistency, and task-driven perceptual loss, with standard ReID objectives such as identity classification, metric learning, and cross-modal contrastive alignment.

\subsection{Baseline Optimization for Extreme Far-Distance Video Person Re-Identification}
\textit{Ashwat Rajbhandari}

\subsubsection{Description of the Algorithm}

This submission addresses the VReID-XFD challenge, which focuses on video-based person re-identification under extreme far-distance conditions characterized by low resolution, large viewpoint variation, and severe domain shifts. The approach builds upon the official DetReIDX baseline and emphasizes systematic optimization of training strategies and hyperparameters, rather than architectural modification.

\paragraph{Method.}
The method adopts the Vision Transformer (ViT-B/16) backbone provided in the baseline, combined with adapter modules and metadata-driven prompt conditioning. Performance improvements are achieved through careful tuning of the training and fine-tuning schedules. Stage-1 training is observed to converge within 30--40 epochs and is finalized at 50 epochs, while Stage-2 fine-tuning converges within 25--30 epochs and is finalized at 40 epochs.

To further enhance performance, k-reciprocal re-ranking is enabled during evaluation, with different neighborhood size ($k$) and weighting ($\lambda$) parameters explored, yielding an improvement of approximately 3--4\% in overall mAP. Training batch sizes are increased to improve identity diversity and feature consistency, while smaller batch sizes are used during evaluation to stabilize feature normalization.

A cosine annealing learning rate schedule is employed to ensure smooth convergence and prevent abrupt optimization instability. Additionally, only a small subset of the final transformer layers is fine-tuned, while the majority of the backbone remains frozen, preserving pretrained representations.

\section{Acknowledgements}

Kailash A. Hambarde work was carried out within the scope of the project “Laboratório Associado”, reference CEECINSTLA/00034/2022, funded by FCT – Fundação para a Ciência e a Tecnologia, under the Scientific Employment Stimulus Program. The author also thanks the Instituto de Telecomunicações for hosting the research and supporting its execution.

Hugo Proença acknowledges funding from FCT/MEC through national funds and co-funded by the FEDER—PT2020 partnership agreement under the projects UIDB/50008/2020 and POCI-01-0247-FEDER-033395.